# Low Data Drug Discovery with One-shot Learning

Han Altae-Tran,[†,§] Bharath Ramsundar,[‡,§] Aneesh S. Pappu,[‡] and Vijay Pande[*,¶]

*Department of Biological Engineering, Massachusetts Institute of Technology, Department of Computer Science, Stanford University, and Department of Chemistry, Stanford University*

E-mail: pande@stanford.edu

## Abstract

Recent advances in machine learning have made significant contributions to drug discovery. Deep neural networks in particular have been demonstrated to provide significant boosts in predictive power when inferring the properties and activities of small-molecule compounds.[1] However, the applicability of these techniques has been limited by the requirement for large amounts of training data. In this work, we demonstrate how one-shot learning can be used to significantly lower the amounts of data required to make meaningful predictions in drug discovery applications. We introduce a new architecture, the residual LSTM embedding, that, when combined with graph convolutional neural networks, significantly improves the ability to learn meaningful distance metrics over small-molecules. We open source all models introduced in this work as part of DeepChem, an open-source framework for deep-learning in drug discovery.[2]

---
[*]To whom correspondence should be addressed
[†]Department of Biological Engineering, Massachusetts Institute of Technology
[‡]Department of Computer Science, Stanford University
[¶]Department of Chemistry, Stanford University
[§]equal contribution

# Introduction

The lead optimization step of drug discovery is fundamentally a low-data problem. When biological studies yield evidence that a particular molecule can modulate essential pathways to achieve therapeutic activity, the discovered molecule often fails as a potential drug for a number of reasons including toxicity, low activity, and low solubility.[3] The central problem of small-molecule based drug-discovery is to optimize the candidate molecule by finding analogue molecules with increased pharmaceutical activity and reduced risks to the patient. Yet, with only a small amount of biological data available on the candidate and related molecules, it is challenging to form accurate predictions for novel compounds.

In the last few years, the field of machine-learning has produced several pivotal advances that address complex problems. Deep neural networks have solved challenging problems in visual perception,[4] speech-recognition,[5] language translation,[6] and game-playing.[7] These techniques leverage the use of multilayer neural network architectures as powerful and flexible function approximators. This capability of deep neural networks is underpinned by their ability to learn sophisticated representations of their input given large amounts of data.

These advances in deep-learning have inspired novel approaches for better understanding chemistry. For example, in 2012, Merck sponsored a Kaggle competition for improving the accuracy of molecular property prediction. The winning team used multitask deep networks and achieved a 15% improvement in relative accuracy over a random forest baseline.[8] Following this work, many groups demonstrated that massively multitask networks can provide significant boosts in the predictive power of deep-learning models for property prediction.[9,10] In parallel, other groups developed sophisticated deep-architectures for processing and extracting features from molecular structures.[11] Graph-convolutional architectures[12,13] in particular process small-molecules as undirected graphs and build up features using learnable convolution layers. In contrast to older small-molecule featurizing algorithms, such as circular fingerprints,[14] these new graph convolutional feature extracting architectures are learnable, meaning they can be modified to improve performance.

The practical effect of these innovations in drug discovery has been limited as most of the aforementioned deep-learning frameworks require large-amounts of data. For example, massively multitask networks have so far been trained with millions of datapoints.[9] This is in stark contrast to the state of current drug discovery pipelines, which often struggle to characterize even a few dozen compounds. Recent work has demonstrated that standard machine-learning techniques such as random forests and simple deep-networks are capable of learning meaningful chemical information from only a few hundred compounds,[15] but even a hundred compounds is often too resource intensive for standard drug discovery campaigns.

Other recent advances in machine learning have demonstrated that in some circumstances, non-trivial predictors may be learned from only a few datapoints.[16–18] These methods work by using related data to learn a meaningful distance metric over the space of possible inputs. This sophisticated metric is used to compare new datapoints to the limited available data and subsequently predict properties of these new datapoints. More broadly, these techniques are known as "one-shot learning" methods. For example, matching-networks,[18] learn a distance-metric upon images which they use to achieve impressive near-human accuracies on the one-shot character-recognition Omniglot dataset.[16]

In this work, we mathematically adapt the standard one-shot learning paradigm to the the drug discovery problem. Standard one-shot learning focuses on recognizing new-classes (say recognizing a giraffe given only one example). In drug-discovery, the challenge is rather to predict the behavior of a molecule in a new experimental system.

We introduce a new deep-learning architecture, the residual LSTM, a modification of the matching-networks architecture and the the residual convolutional network.[19] This architecture allows for the learning of sophisticated metrics which can trade information between evidence and query molecules. We demonstrate that this architecture offers significant boosts in predictive power for a variety of problems meaningful for low-data drug discovery.

Furthermore, we take a strong open-source approach in this work. Every primitive introduced in this paper is open-sourced as part of the DeepChem[2] library for open-source drug discovery. In particular we provide high-quality Tensorflow[20] implementations of graph-convolutional primitives along with implementations of our one-shot learning models. The scripts used to generate all numbers in this paper are similarly open sourced, along with all datasets. Interested parties can reproduce all results claimed in this work with relative ease.

# Methods

## Mathematical Formalism

In this paper, we consider the situation in which one has multiple binary learning tasks. Some proportion of these tasks are reserved for training a one-shot learning model. The remaining tasks are those with too little data for standard machine-learning models to learn an effective classifier predicting the outcomes for these tasks (active/inactive correspond respectively to positive/negative examples for the binary classifier). The goal is to harness the information available in the training tasks to create strong classifiers for the test systems.

Mathematically, we have $T$ tasks, each associated with a data set, $S = \{(x_i, y_i)\}_{i=1}^m$, $y_i \in \{0, 1\}$. We refer to the collection of available datapoints for a given task as a "support" set. The goal is to learn a function $h$, parameterized upon choice of support $S$ that predicts the probability of any query $x$ being active in the same system. Formally, $h_S(x) : \mathcal{X} \to [0, 1]$, where $\mathcal{X}$ is the chemical space of small-molecules. If $h$ is specified with a neural network, fully differentiable with respect to $x$ and $S$, then $h$ can be trained end-to-end with gradient descent.

## Simple One-shot learning

Deep one-shot learning systems[17,18] use convolutional layers to encode images into continuous vectors. In the simplest one-shot learning formalism these continuous vectors are then fed into a simple nearest-neighbor classifier, that labels new examples by distance-weighted combination of support set labels. Let $a(x, x_i)$ denote some weighting function for query example $x$ and support set element $x_i$ with associated label $y_i$. Then the label $h_S(x)$ for query compound $x$ can be defined as

$$h_S(x) = \sum_{i=1}^{m} a(x, x_i) y_i$$

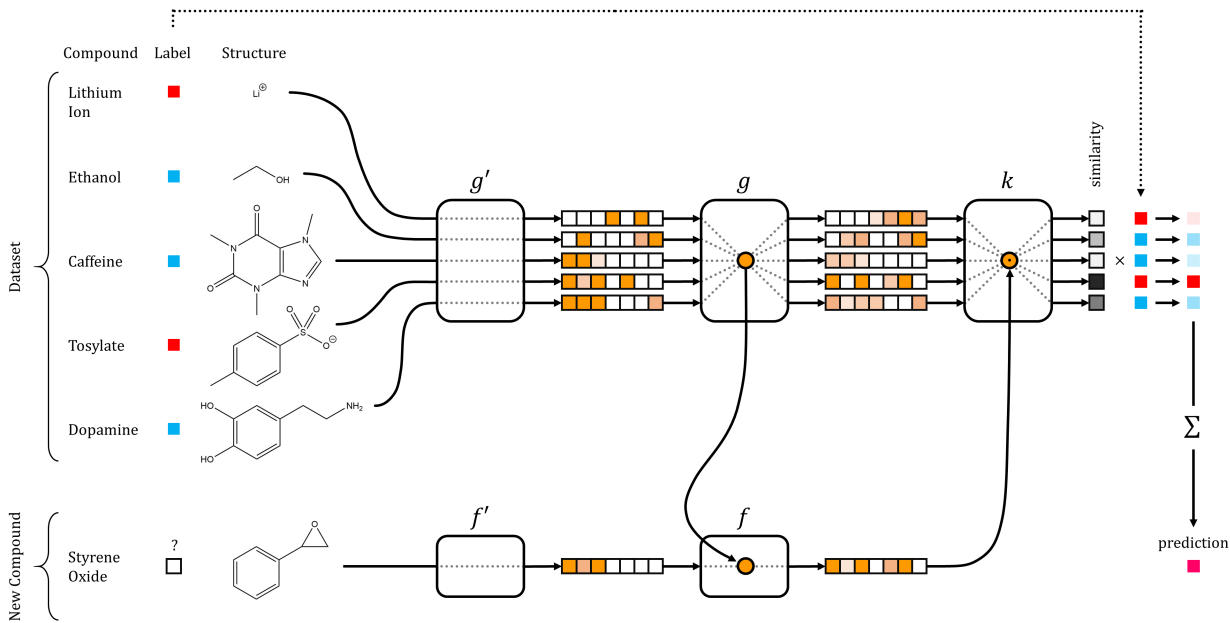

Figure 1: Schematic of Network Architecture for one-shot learning in drug discovery

The function $a$ is called an attention mechanism in previous work.[18] Mathematically, an attention mechanism is a non-negative function $a(x, x_i)$, with $\sum_{i=1}^{m} a(x, x_i) = 1$, that forms a probability distribution over the support set. The purpose of the attention distribution $a(x, \cdot)$ is to reflect the model's reasoning about the relationship between examples in support $S$ and the query $x$. The final prediction of the task's output for $x$ can then be interpreted as the expected value of the $y_i$'s under the attention distribution.

The attention mechanism $a$ depends on two embedding functions $f : \mathcal{X} \to \mathbf{R}^p$ and $g : \mathcal{X} \to \mathbf{R}^p$ which embed input examples (in small molecule space) into a continuous representation space. In past work, both $f$ and $g$ are standard convolutional networks, while in ours, $f$ and $g$ are graph-convolutional networks. The similarities between the "embed-ded" vectors $f(x)$ and $\{g(x_i)\}$ are computed using a similarity measure $k(\cdot, \cdot)$. For example, $k$ could be the cosine-distance between two vectors. Given a choice of similarity measure, a simple attention mechanism $a$ can be defined by normalizing the similarity values

$$a(x, x_i) = \frac{k\left(f(x), g(x_i)\right)}{\sum_{j=1}^{m} k\left(f(x), g(x_j)\right)}$$

Note that $a(x, x_i)$ is large when $f(x)$ is closer to $g(x_i)$ under metric $k$ compared to the other $g(x_j)$'s. This formulation of one-shot learning has been referred to as a Siamese one-shot learning method.[21]

## Attention Based Embedding

The simple one-shot learning formulation in the previous section uses independent embeddings $f(x)$ and $g(x_i)$ with only the distance metric $k$ linking the information from the two feature maps together. As a result, the feature map $f(x)$ is formed without any context about data available in support $S$. The previously introduced matching-network architecture[18] addresses this problem by developing full context embeddings, in which embeddings $f(x) = f(x \mid S)$ and $g(x_i) = g(x_i \mid S)$ are computed using both $x$ and $S$. Full context embeddings allow the embeddings for $x$ and $x_i$'s to affect one another. Empirically, this addition allows for stronger one-shot learning results.

To construct $f(x \mid S)$ and $g(x_i \mid S)$, matching networks[18] compute embeddings $f'(x)$ and $g'(x)$ using standard convolutional neural networks. The embedding $g(x)$ is computed by placing all the $g'(x_i)$'s into a vector, and linking all the elements using a bidirectional LSTM[22,23] (BiLSTM), allowing each $g(x_i)$ to be influenced by all the $g'(x_j)$'s.

$$g(x \mid S) = \text{BiLSTM}([g'(x_1)|\cdots|g'(x_m)])$$

To compute embedding $f(x \mid S)$, matching networks exploit a context based attention LSTM

model[24] (attLSTM). This model computes an order independent combination of its inputs.

$$f(x \mid S) = \text{attLSTM}(f'(x), \{g(x_i \mid S)\})$$

Both the BiLSTM and attLSTM are mechanisms for combining sequences of vectors into single vectors. However, the attLSTM is order-independent, while the BiLSTM is order dependent. The order dependence in the definition of $g(x \mid S)$ is undesirable since there is no natural ordering to the elements of the support set. Furthermore, the treatment of $f$ and $g$ is non-symmetric. While $g(\cdot \mid S)$ depends only on the $g'$, the definition of $f$ depends on $f'$ and the embedding $g(\cdot \mid S)$.

## Iterative refinement using Dual Residual LSTMs

Our proposed architecture for one-shot learning preserves the context-aware design of matching networks, but resolves the order dependence in the support-embedding $g$ and the non-symmetric treatment of the query and support noted in the previous section.

The core idea is to use an attLSTM to generate both query embedding $f$ and support embedding $g$. As noted previously, the matching networks[18] embedding requires the support embedding $g(\cdot \mid S)$ to define $f(\cdot \mid S)$. To resolve this issue, our architecture iteratively evolves both embeddings simultaneously.

On the first iteration, we define $f(x) = f'(x)$ and $g(S) = g'(S)$. Then, we iteratively update the embeddings $f$ and $g$ using an attention mechanism. This construction allows the embeddings to iteratively inform one another. The equations below specify the full update performed in each iteration.

$$\textit{Initialize}$$
$$\mathbf{r} = g'(S) \quad \delta\mathbf{z} = \mathbf{0} \quad \delta z = 0$$
$$\textit{Repeat}$$
$$e = k(f'(x) + \delta z, \mathbf{r}) \qquad \mathbf{e} = k(\mathbf{r} + \delta\mathbf{z}, g'(S)) \quad \text{(similarity measures)}$$
$$a_j = e_j / \sum_{j=1}^m e_{ij} \qquad \mathbf{A}_{ij} = \mathbf{e}_{ij} / \sum_{j=1}^m \mathbf{e}_{ij} \quad \text{(attention mechanism)}$$
$$r = a^\top \mathbf{r} \qquad \mathbf{r} = \mathbf{A} g'(\mathbf{S}) \quad \text{(expected feature map)}$$
$$\delta z = \text{LSTM}\left([\delta z, r]\right) \qquad \delta\mathbf{z} = \text{LSTM}\left([\delta\mathbf{z}, \mathbf{r}]\right) \quad \text{(generate updates)}$$
$$\textit{Return}$$
$$f(x) = f'(x) + \delta z \qquad g(\mathbf{S}) = g'(\mathbf{S}) + \delta\mathbf{z} \quad \text{(evolve embeddings)}$$

The generated dual embeddings are used to perform one-shot learning as in the simpler models explained in previous sections.

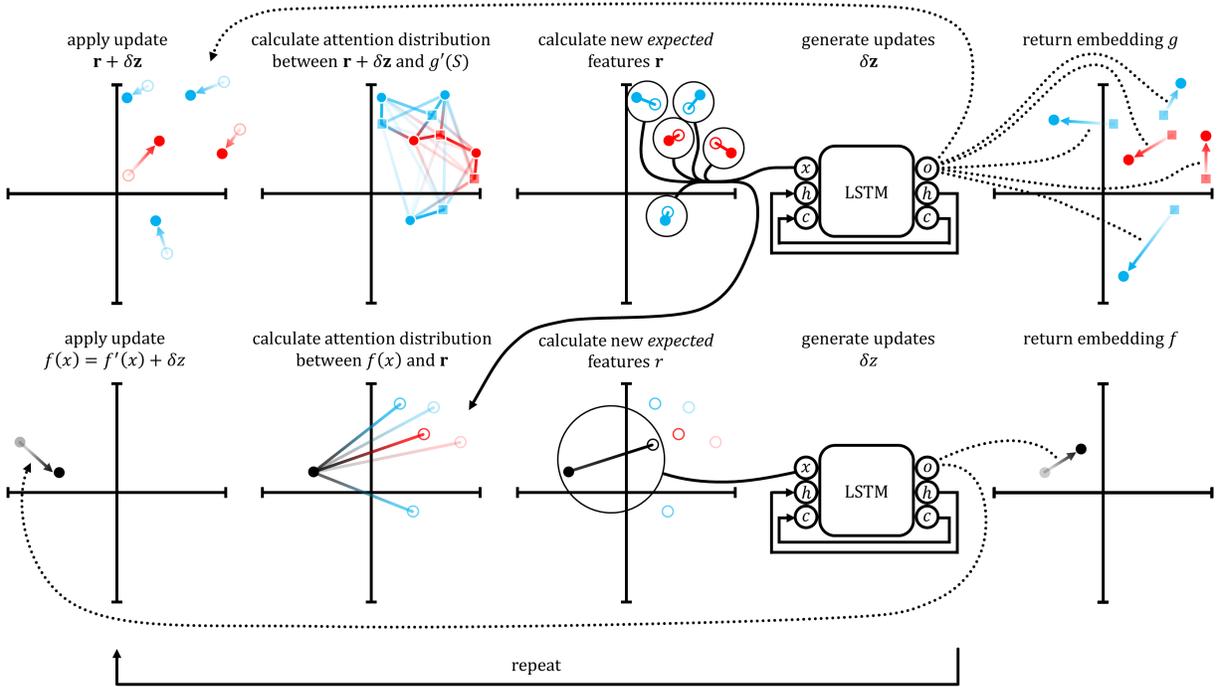

Figure 2: Pictorial depiction of iterative refinement of embeddings. Red and blue points depict positive/negative samples (for illustrative purposes only). The original embedding $g'(S)$ is shown as squares. The expected features, $\mathbf{r}$ are shown as empty circles.

## Graph Convolutions

Earlier work[12] defines a generalized graph convolution layer that extends standard two-dimensional convolutions upon images to arbitrary graphs. In standard convolutional networks, an image can be viewed as a grid, where each node corresponds to a pixel. The "neighbors" of a pixel in the same receptive field are passed through a dense neural layer to compute the output value for the convolution.[25] Similarly, when computing the convolution output for a specific node in an arbitrary graph, all neighbors of the node are passed through a dense layer to form the new features at this node. Both convolutions exploit the "local geometry" of the system to reduce the number of learnable parameters (see figure 3 and Appendix).

Standard convolutional networks have "pooling layers", which perform a max operation upon local receptive fields in an image.[25] In analogy to this pooling operation, we introduce an analogous max pool operator on a node in graph structured data that simply returns the maximum activation across the node and the node's neighbors (see figure 3 and Appendix). Such an operation is useful for increasing the size of downstream convolutional layer receptive fields without increasing the number of parameters.

In a graph-convolutional system, each node has a vector of descriptors. However, at prediction time, we will require a single vector descriptor of fixed size for the entire graph. We introduce a new graph-gather convolutional layer which simply sums all feature vectors for all nodes in the graph to obtain a graph feature vector (see figure 3).

These graph-convolutional operations are useful for transforming small-molecules into continuous vectorial representations. Such representations are emerging as a powerful way to manipulate small molecules within deep networks.[26] Earlier work on one-shot learning uses convolutional neural networks to encode images into continuous vectors which are then used to make one-shot predictions. By analogy, we use graph-convolutional neural networks to encode small-molecules in a form suitable for one-shot prediction.

To facilitate the use of graph-convolutional layers in future work, we open-source our Ten-sorflow[20] implementation of graph-convolutions, graph-pooling, and graph-gathering layers as part of DeepChem.[2]

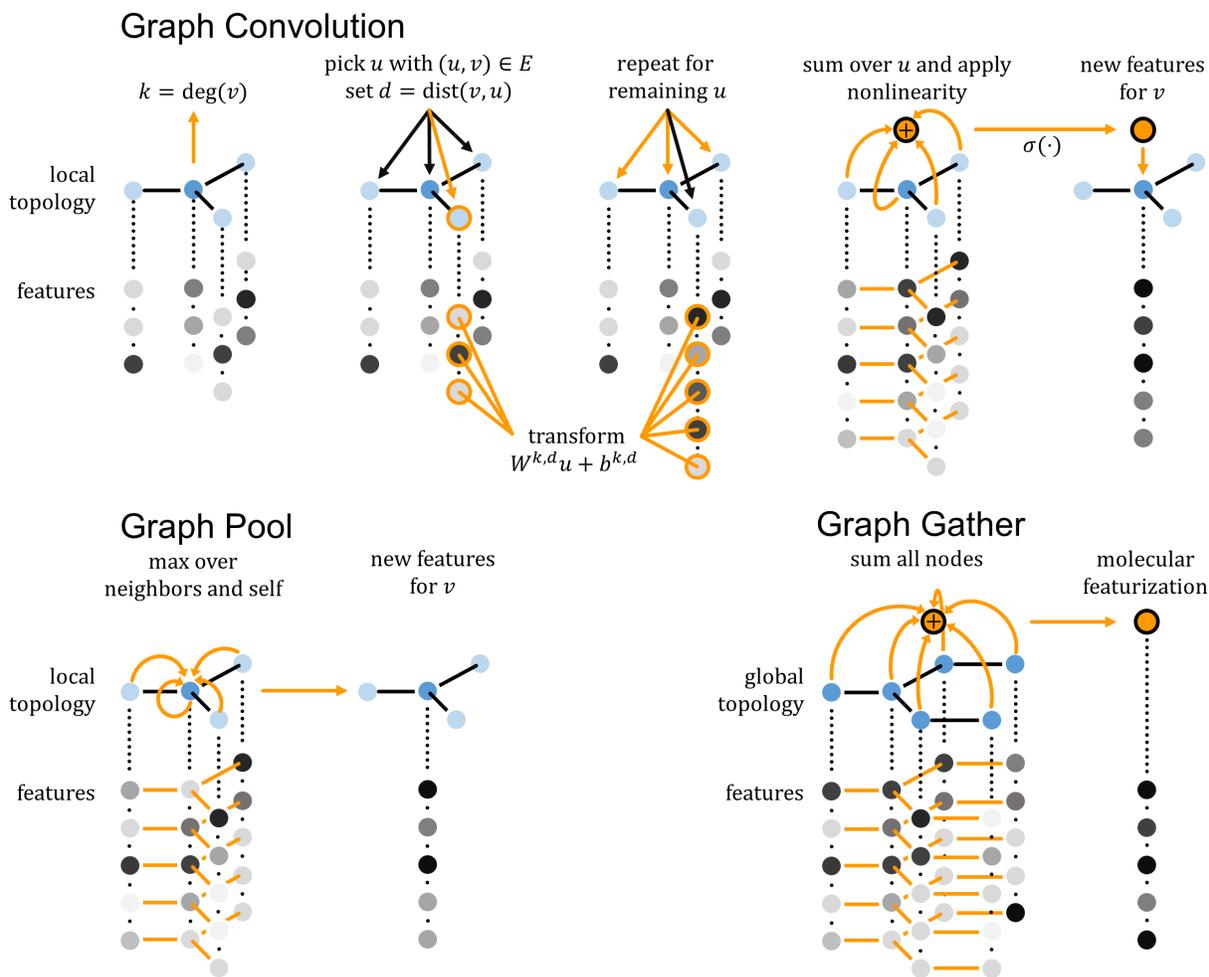

Figure 3: Graphical representation of the major graph operations described in this paper. For each of the operations, the nodes being operated on are shown in blue, with unchanged nodes shown in light blue. For graph convolution and graph pool, the operation is shown for a single node, $v$, however, these operations are performed on all nodes $v$ in the graph simultaneously.

## Model Training

The objective for the Residual LSTM models is similar to that for the matching networks, with the key difference that a set of training tasks are specified instead of a set of training classes. Let Tasks represent the set of training tasks. Let $S$ represent a support set, and let $B$ represent a batch of queries.

$$\theta = \arg\max_{\theta} E_{T \in \text{Tasks}} \left[ E_{S \sim T,, B \sim T} \left[ \sum_{(x,y) \in B} \log P_\theta(y \mid x, S) \right] \right]$$

Training consisted of a sequence of episodes. In each episode, a task $T$ was randomly sampled, then a support $S$ and a batch of queries $B$ (non-overlapping) were sampled from the task. One gradient descent step, using ADAM[27] was performed for each episode. We experimented with more gradient descent steps per episode, but found that sampling more supports instead improved performance.

# Experiments

### Tox21

The Tox21 collection consists of 12 nuclear receptor assays related to human toxicity, first gathered for the Tox21 Data Challenge.[28] The winner of this challenge used a multitask deep-network[29] to achieve strong predictive performance.

For a low-data benchmark, we don't use the standard train/test split for this dataset. Instead, we use the first nine assays as the training set, and the last 3 assays as the test collection. For details, see appendix.

Results are in Table 1. All one-shot learning methods show strong boosts over the random-forest baseline, with the Residual LSTM models showing a more robust boost in the presence of less data.

### SIDER

The SIDER dataset contains information on marketed medicines and their observed adverse reactions.[30] We originally tested on the original 5868 side effect categories, but due to lack of signal at this level of granularity we grouped the drug-SE pairs into 27 system organ classes according to MedDRA classifications.[31] We treat the problem of predicting whether a given compound induces a side-effect as an individual task (for 27 tasks in total). For the low-data benchmark, all models were trained on the first 21 tasks and tested on the last 6. For details see appendix.

Table 1: Accuracies of models on held-out tasks for Tox21. Numbers reported are median on test-tasks. Numbers for each task are averaged for 20 random choices of support sets.

| Tox21 | RF (50 trees) | RF (100 trees) | Siamese | AttnLSTM | ResLSTM |
|---|---|---|---|---|---|
| 10 pos, 10 neg | 0.537 | 0.563 | **0.831** | **0.834** | **0.840** |
| 5 pos, 10 neg | 0.537 | 0.579 | 0.790 | **0.820** | **0.837** |
| 1 pos, 10 neg | 0.537 | 0.584 | 0.710 | 0.687 | **0.757** |
| 1 pos, 5 neg | 0.571 | 0.572 | 0.689 | 0.595 | **0.815** |
| 1 pos, 1 neg | 0.536 | 0.542 | 0.668 | 0.652 | **0.784** |

Results are in Table 2. All one-shot learning methods show strong boosts over the random-forest baseline, with the Residual LSTM models showing a more robust boost in the presence of less data.

Table 2: Accuracies of models on held-out tasks for SIDER. Numbers reported are median on test-tasks. Numbers for each task are averaged for 20 random choices of support sets

| SIDER | RF (50 trees) | RF (100 trees) | Siamese | AttnLSTM | ResLSTM |
|---|---|---|---|---|---|
| 10 pos, 10 neg | 0.551 | 0.546 | 0.660 | 0.671 | **0.752** |
| 5 pos, 10 neg | 0.534 | 0.541 | 0.674 | 0.671 | **0.750** |
| 1 pos, 10 neg | 0.537 | 0.533 | 0.542 | 0.543 | **0.602** |
| 1 pos, 5 neg | 0.536 | 0.535 | 0.544 | 0.539 | **0.639** |
| 1 pos, 1 neg | 0.504 | 0.501 | 0.506 | 0.505 | **0.623** |

## MUV

The MUV dataset collection[32] contains 17 assays designed to be challenging for standard virtual screening. The positives examples in these datasets are selected to be structurally distinct from one another. As a result, this collection is a best-case scenario for baseline machine learning (since each data point is maximally informative) and a worst-case test for the low-data methods, since structural similarity cannot be effectively exploited to predict behavior of new active compounds.

Results are in Table 3. The random forest baseline does significantly better for MUV than for the other two datsets we tested. Though the residual LSTM does well, it does not achieve the best performance. This result suggests that one-shot learning methods may have some difficulties generalizing to new molecular scaffolds.

Table 3: Accuracies of models on held-out tasks for MUV. Numbers reported are median on test-tasks. Numbers for each task are averaged for 20 random choices of support sets

| SIDER | RF (50 trees) | RF (100 trees) | Siamese | AttnLSTM | ResLSTM |
|---|---|---|---|---|---|
| 10 pos, 10 neg | 0.710 | **0.741** | 0.501 | 0.683 | **0.712** |
| 5 pos, 10 neg | **0.723** | **.751** | 0.708 | 0.674 | 0.672 |
| 1 pos, 10 neg | 0.586 | **0.624** | 0.567 | 0.583 | **0.619** |
| 1 pos, 5 neg | 0.561 | 0.579 | 0.546 | 0.565 | **0.634** |
| 1 pos, 1 neg | 0.558 | 0.573 | 0.498 | 0.501 | 0.512 |

## Transfer Learning to SIDER from Tox21

The experiments thus far have tested the ability of one-shot learning methods to generalize from training tasks to closely related testing tasks. To test whether one-shot learning methods are capable of broader generalization, we trained models on the Tox21 dataset collection and evaluated predictive accuracy on the SIDER collection. Note that these collections are broadly distinct, with Tox21 measuring results in nuclear receptor assays, and SIDER measuring adverse reactions from real patients.

Results are in Table 4. None of our models achieve any predictive power on this task (the random forest numbers are copied over from Table 2 for comparison), providing evidence that the one-shot models have limited generalization powers to unrelated systems.

Table 4: Accuracies of models trained on Tox21 and evaluated on SIDER. Random forest numbers copied over from SIDER table for comparative purposes. Numbers reported are median on SIDER tasks. Numbers for each task are averaged for 20 random choices of support sets

| SIDER from Tox21 | RF (50 trees) | RF (100 trees) | Siamese | AttnLSTM | ResLSTM |
|---|---|---|---|---|---|
| 10 pos, 10 neg | .551 | .546 | 0.504 | 0.510 | .509 |

# Discussion and Conclusion

This paper introduces the task of low data learning for drug discovery, and provides an architecture for learning such models. We demonstrate that this architecture can provide strong boosts over simpler methods for low-data learning. On the Tox21 and SIDER collections, one-shot learning methods strongly dominate simpler machine learning baselines. This result is particularly interesting for the SIDER collection, since this dataset consists of very high-level phenotypic side-effect observations. Given the amount of uncertainty in these predictions, the fact that one-shot learning is able to do well is a strong indication that these methods could offer strong performance on small biological datasets (for example, perhaps on a small number of drug tests done with rats).

At the same time, it is clear that there are strong limitations to the generalization powers of current one-shot learning methods. On the MUV datasets, one-shot learning methods struggle (but hold their own) against simple machine-learning baselines. This result is likely due to the presence of diverse scaffolds in the MUV collection, suggesting that one-shot learning methods may struggle to generalize to novel molecular scaffolds. Furthermore, on the transfer learning experiment, which attempts to use the Tox21 collection to train SIDER predictors, all one-shot learning methods collapse entirely. It is clear that there is a limit to the cross-task generalization capability of one-shot models, but it is left to future work to determine the precise limits.

In order to facilitate the wide adoption of these models, we've open-sourced all graph-convolutional primitives in addition to the Residual LSTM models themselves as part of the DeepChem library. All scripts used to perform experiments listed in this paper have been made public as well.

The use of one-shot learning in chemistry can only be validated experimentally, but we hope that our results will the impetus for such work.

**Acknowledgments**

We would like to thank the Stanford Computing Resources for providing us with access to the Sherlock and Xstream GPU nodes. Thanks to David Duvenaud for useful preliminary discussions.

B.R. was supported by the Fannie and John Hertz Foundation.

# Appendix

### Definitions of Graph Primitives

There are three major neural-network layers that are used to featurizing the molecular graphs. This is the graph convolution, $h_{\text{conv}}(G)$, the graph pool, $h_{\text{pool}}(G)$, and the graph gather, $h_{\text{gather}}(G)$, all defined below. The graph convolution and graph pool operations definitions are given for a single node in the graph $v \in V$; however, when performing the operation on the graph, $G$, the operation is performed on all nodes simultaneously. This means that $h_{\text{conv}}(G) = [h_{\text{conv}}(v_1), h_{\text{conv}}(v_2), \ldots]$, and similarly for the pool layer. Specifically, we define

$$h_{\text{conv}}(v) = \sigma\left(\sum_{(u,v) \in E} W^{\deg(v)} v + U^{\deg(v)} u + b^{\deg(v)}\right)$$

$$h_{\text{pool}}(v) = \max\left\{\max_{(u,v) \in E} u, v\right\}$$

$$h_{\text{gather}}(G) = \sum_{u \in V} u$$

where $\sigma(\cdot)$ is a nonlinearity, such as ReLU or tanh.

## Convolutional Architecture in this Paper

The Siamese, AttnLSTM, and ResLSTM models all used the same convolutional architecture, shown in the table below, with the input starting at the left, sequentially feeding into the layers to the right.

Table 5: Convolutional Network Architecture

| layer | conv | pool | conv | pool | conv | pool | dense | gather |
|---|---|---|---|---|---|---|---|---|
| dimension | 64 | | 128 | | 64 | | 128 | |
| nonlinarity | relu | | relu | | relu | | tanh | tanh |

## Molecular Features

For the convolutional models, all molecules were featurized into graphs by considering atoms as nodes and bonds as edges in an undirected graph. No distinction was made between bond types. RDKit[33] was used to compute basic features of atoms including atom-type, valences, formal charges, and hybridization for each atom in a given molecule. This set of features formed the initial set of atomic features fed into graph-convolutional layers.

## Tox21 Details

The assays NR-AR, NR-AR-LBD, NR-AhR, NR-Aromatase, NR-ER, NR-ER-LBD, NR-PPAR-gamma, SR-ARE, SR-ATAD5 were used for training. Assays SR-HSE, SR-MMP, and SR-p53 were used for model evaluation.

## SIDER Details

Indications "Hepatobiliary disorders", "Metabolism and nutrition disorders", "Product issues", "Eye disorders", "Investigations,Musculoskeletal and connective tissue disorders",

"Gastrointestinal disorders", "Social circumstances", "Immune system disorders", "Reproductive system and breast disorders", "Neoplasms benign, malignant and unspecified (incl cysts and polyps)", "General disorders and administration site conditions", "Endocrine disorders", "Surgical and medical procedures", "Vascular disorders", "Blood and lymphatic system disorders", "Skin and subcutaneous tissue disorders", "Congenital, familial and genetic disorders", "Infections and infestations","Respiratory, thoracic and mediastinal disorders", "Psychiatric disorders" were used for training. Indications "Renal and urinary disorders", "Pregnancy, puerperium and perinatal conditions", "Ear and labyrinth disorders", "Cardiac disorders", "Nervous system disorders", "Injury, poisoning and procedural complications" were used for model evaluation.

## MUV Details

Assays MUV-466, MUV-548, MUV-600, MUV-644, MUV-652, MUV-689, MUV-692, MUV-712, MUV-713, MUV-733, MUV-737, MUV-810 were used for training. Assays MUV-832, MUV-846, MUV-852, MUV-858, MUV-859 were used for model evaluation.